\pdfoutput=1  % REQUIRED by arXiv: forces pdflatex, needed because figures are .png

\documentclass[11pt]{article}

\usepackage{preprint}

\usepackage[T1]{fontenc}
\usepackage[utf8]{inputenc}
\usepackage{lmodern}
\usepackage{microtype}
\usepackage{amsmath}
\usepackage{graphicx}
\usepackage{booktabs}
\usepackage{multirow}
\usepackage{array}
\usepackage{enumitem}
\usepackage{tcolorbox}
\usepackage[font=small,labelfont=bf,labelsep=period]{caption}
\usepackage[round,authoryear]{natbib}
\usepackage{url}
\PassOptionsToPackage{colorlinks=true,allcolors=black,citecolor=black,urlcolor=black}{hyperref}
\usepackage{hyperref}
\usepackage{orcidlink}

\tcbuselibrary{skins,breakable}
\defcitealias{analyzeboston}{Analyze Boston, n.d.}

\newcommand{\vulnprompt}[1]{%
  \begin{tcolorbox}[breakable,enhanced,colback=black!3,colframe=black!35,
    boxrule=0.4pt,left=9pt,right=9pt,top=8pt,bottom=8pt,arc=1pt]
  \small
  \setlength{\parskip}{0.6em}\setlength{\parindent}{0pt}%
  #1
  \end{tcolorbox}}

\newcolumntype{R}{>{\raggedleft\arraybackslash}X}

\title{Using Fine-Tuned LLMs to Identify Indicators of\\Vulnerability in UK Police Incident Logs}

\newcommand{\authorblock}{%
  \raggedright
  \textbf{Sam Relins}\textsuperscript{a,b}\,\orcidlink{0009-0001-7868-9835}\quad
  \textbf{Daniel Birks}\textsuperscript{a,b,*}\,\orcidlink{0000-0003-3055-7398}\par
  \vspace{0.8em}
  {\small
   \textsuperscript{a}\,ESRC Vulnerability and Policing Futures Research Centre\\
   \textsuperscript{b}\,School of Law, University of Leeds, Leeds, UK\\[0.5em]
   \textsuperscript{*}\,Corresponding author:
   \href{mailto:d.birks@leeds.ac.uk}{d.birks@leeds.ac.uk}\par}%
}
\author{\authorblock}

\headertitle{Fine-Tuned LLMs and Vulnerability in UK Police Incident Logs}

\begin{document}

\maketitle

\begin{abstract}
\textbf{Purpose.} Understanding how much of routine policing involves vulnerable people could
inform resourcing, training, and multi-agency response, yet administrative data provide limited
insight. We explore whether an LLM-based classification pipeline, developed on open-source US
police data, can be adapted to estimate the prevalence of four vulnerability indicators - mental
ill health, substance misuse, alcohol dependence, and homelessness - in UK police incident
narratives, and when outputs can be treated as defensible measurements.

\textbf{Methods.} We analyse nearly 3,000 de-identified incident logs from a UK police force,
using a multi-stage pipeline combining repeated model inference, label aggregation, structured
human review, and statistical correction. The pipeline runs on a locally hosted open-weight LLM,
reflecting secure compute environments police are likely to use. Our approach assesses output
stability, quantifies biases, and generates corrected estimates with uncertainty.

\textbf{Results.} LLMs can produce meaningful, if imperfect, prevalence estimates at scale.
Mental ill health indicators are present in approximately one in five incidents, with lower
prevalence for other indicators. However, na\"ive LLM deployment is unreliable: single-pass
classifications are unstable, and aggregated outputs systematically over-assign indicators
relative to human judgement. Correcting these biases required substantial human input and
statistical adjustment, leaving considerable uncertainty.

\textbf{Conclusions.} While LLMs can extract information from unstructured police data, their
outputs cannot be treated as valid measurements without careful methodological support. At the
population level, defensible estimates are achievable but resource-intensive; at the individual
level, errors remain frequent and unpredictable, limiting suitability for operational decisions.
This study highlights the potential and constraints of LLM-based measurement, underscoring the
need for rigorous evaluation in applied settings.
\end{abstract}


\newpage

% =============================================================== INTRODUCTION
\section{Introduction}

Frontline policing routinely brings officers into contact with people experiencing complex and
overlapping forms of need. These types of contact - widely grouped under the term
vulnerability - are well documented in England and Wales and across comparable international
jurisdictions \citep{crawford2024}. Mental ill health, drug abuse, alcohol dependence, and
homelessness feature prominently in the incidents that frontline officers attend and the
cumulative demand associated with these encounters is widely understood to be substantial,
placing sustained pressure on policing resources and shaping patterns of police activity.
Estimates vary - officers themselves report that between 40\% and 75\% of their work involves
vulnerable individuals \citep{met_npcc}, while structured data analyses tend to produce lower but
still significant figures \citep{kane2021, haas2024}. This has prompted policy responses such as
Right Care, Right Person, whose national rollout reflects official recognition of both the strain
placed on policing capacity and the imperfect alignment between police roles and the complex
needs encountered \citep{homeoffice2024, cqc2025}. It also raises important
questions about capacity and coordination across health and wider public services.

Accurately estimating the prevalence of vulnerability-related interactions in policing is
desirable for informing both frontline resource allocation and broader strategic policy
responses. At a force level, it has the potential to inform resourcing decisions, training
priorities, and the design of partnerships with health and social care agencies. At a policy
level, it could support robust, evidence-informed debates about the appropriate boundaries of the
police role, the adequacy of mental health provision, and the allocation of public funding across
services that share responsibility for vulnerable populations. Currently, the evidence base for
these conversations remains remarkably thin. Structured administrative data - the qualifier
flags, incident categories, and disposal codes that police information systems record - capture
some of this picture, but they were understandably designed for operational and administrative
purposes, rather than an epidemiological assessment of vulnerability-related demand. They are
known to be incomplete, inconsistently applied, and shaped by recording practices that vary
between forces and over time \citep{haas2024, hmicfrs2018}. As a result, it is difficult to say
with any confidence how often police encounter specific vulnerabilities in their day-to-day work.

What police forces do possess, in enormous volume, is narrative text. Incident logs,
command-and-control records, and officer reports contain unstructured free-text descriptions of
police--public interactions that are often far richer than the structured fields that accompany
them. These narratives routinely describe the circumstances, behaviours, and stated needs of the
individuals involved. However, this information has historically been inaccessible at scale.
Manual qualitative coding of thousands of incident narratives is prohibitively resource-intensive,
and conventional natural language processing methods, while productive for tasks such as entity
extraction and topic modelling in policing contexts
\citep[see][for a comprehensive review]{lukmanjaya2026}, have lacked the flexibility to perform the
kind of interpretive, context-sensitive classification that vulnerability identification demands.

Recent advances in large language models (LLMs) have substantially changed this picture.
Instruction-tuned models can follow natural-language task descriptions, evaluate ambiguous
evidence in context, and produce classifications with accompanying justifications - capabilities
that align well with the requirements of deductive coding in qualitative research. A growing body
of work has begun to explore LLM applications in criminal justice settings, from report writing
assistance \citep{adams2024} to crime classification \citep{bifari2024}, though empirical
evaluations of their analytical capabilities remain limited. In a prior study
\citep{relins2025}, we assessed whether instruction-tuned LLMs could replicate human qualitative
coding of vulnerability indicators in publicly available narrative reports from the Boston Police
Department, testing multiple models, prompting strategies, and assessing potential demographic
biases through counterfactual analyses across four vulnerability types: mental ill health,
substance misuse, alcohol dependence, and homelessness. The primary finding of that study
demonstrated that LLMs are highly effective at identifying narratives where vulnerabilities are
absent; functioning as reliable negative filters, while also highlighting significant limitations
in positive and ambiguous classifications, and the continued need for human oversight.

The present study extends this approach to sensitive operational data provided by a UK police
force. Using three months of incident logs from a single basic command unit located within a UK
police force, we explore whether the classification method developed on
public US data can be adapted to operate within the practical and ethical constraints of highly
sensitive policing data. We further assess whether the resulting classifications, once subjected
to human validation and statistical adjustment, can generate something that structured police
data currently cannot: robust estimates of the prevalence of vulnerability-related interactions,
specifically those involving mental ill health, substance misuse, alcohol dependence, and
homelessness.

In what follows, we provide a detailed account of the methodological challenges encountered - the
model's behaviour during classification, the adaptations required when that behaviour deviated
from expectations, and the statistical corrections necessary to produce estimates with an
acceptable degree of confidence. The study contributes both a set of novel prevalence estimates
and a practical demonstration of what responsible LLM-based measurement entails in high-stakes
applied settings. The former adding to a growing evidence base on joined-up service provision;
the latter advancing the integration of evidence-based practices with the responsible development
and deployment of large language models in applied policing contexts.

% ==================================================================== METHOD
\section{Method}

\subsection{Data}

Three months of STORM incident logs - records generated by the force's computer-aided dispatch
system - were obtained from a non-metropolitan UK police force. STORM is a computer-aided dispatch
system widely used across British policing; incident logs are created and updated by control room
operators and attending officers as events unfold, recording details such as the nature of the
call, the location, individuals involved, and actions taken. Because entries are entered in real
time under operational pressure, logs are frequently unstructured and idiosyncratic - officers and
operators develop shorthand conventions, entries vary considerably in length and completeness, and
the same incident may accumulate multiple overlapping updates from different contributors. The
logs also routinely contain personal information: names, addresses, telephone numbers, and
references to third parties are embedded within free-text fields as a matter of course, reflecting
the operational purpose of the system rather than any expectation of subsequent research use.

The analysed logs related to a single Basic Command Unit (BCU), the largest territorial
subdivision of a British police force, within a non-metropolitan UK force, covering a three-month
period and totalling just under 3,000 individual logs. The specific force is not identified here
and the BCU and time period were not disclosed to the research team; it is known only that the
data were drawn from a BCU containing both urban and rural areas and were extracted at some point
between 2022 and 2025, and would therefore not be directly affected by pandemic-related
restrictions.

To protect against disclosure within the logs themselves, the data were anonymised in-force using
a whitelisting approach. Force analysts were provided with a purpose-built tool that generated a
vocabulary of all words appearing in the corpus; the most frequently occurring terms were then
manually verified on-site by researchers to confirm they did not constitute personal identifiers.
This approved whitelist was subsequently used to filter all logs, with any word not appearing on
it redacted prior to transfer. Of the 1,181,029 word tokens in the corpus, 81,098 (6.9\%) were
redacted prior to transfer, with 1,099,931 (93.1\%) retained. The processed data were subsequently
stored and processed within a university secure research environment.

Prior to classification, raw logs were cleaned to remove entries with no narrative value,
including system-generated messages, disposal records, and automated tags. Individual log entries
belonging to the same incident were stitched together into a single narrative per incident based
on a unique incident ID. Where a stitched narrative exceeded 400 words, it was divided into
smaller chunks for classification. This threshold was determined by the operational constraints of
the secure research environment: because data of this sensitivity could not be processed using
external APIs or cloud-based infrastructure, classification was performed using a locally-hosted
model with a comparatively limited context window. A 400-word limit provided a conservative margin
within that window, accommodating both the incident narrative and the classification prompt
without risk of truncation. Chunks were created at natural breakpoints - paragraph boundaries,
sentence endings, and clause boundaries, in that order of preference - to preserve narrative
coherence and avoid splitting text at points that might obscure meaning for the model. Where an
incident was divided into multiple chunks, each was classified independently; the aggregation of
chunk-level classifications into a single incident-level outcome is described below. These
pre-processing steps produced a final corpus of 3,850 text segments from 2,977 incident logs.

\subsection{Two-Step Classification}
\label{sec:twostep}

We use a two-step classification approach, adapted from our earlier work classifying Boston Police
FIO reports \citep{relins2025}. That study established two findings that motivate this design:
(i) LLMs perform strongly as negative classifiers - reliably identifying where there is no
evidence of a given vulnerability, and (ii) the majority of incidents contain no evidence of
vulnerability indicators. An initial screening step therefore identifies logs that can confidently
be classified as negative, with only those passing this threshold proceeding to substantive
classification. The method was applied separately for each of four vulnerability types: mental ill
health, substance misuse, alcohol dependence, and homelessness. The prompts shown below use mental
ill health as the example, which follows the two-step structure in its most straightforward form.

In the first step, the model is presented with an incident narrative and asked to highlight any
quotes from the text that may indicate the vulnerability in question. To optimise this process,
the prompt explicitly limits the model to extracting a maximum of two quotes. During initial
experiments, we observed that without this constraint, the model tended toward verbosity,
utilising excessive tokens to highlight marginal or spurious evidence - a particular concern given
the limited context window of the locally-hosted model. Restricting extraction to two quotes
effectively curbed this, improving computational efficiency while maintaining the same level of
downstream classification accuracy. If no indicators are found, the model responds with
``EVIDENCE: NONE'' and the classification process ends. This substantially reduces processing for
the majority of cases where no evidence is present.

\vulnprompt{%
\textbf{System:} \emph{You are an assistant tasked with identifying statements in incident reports
that might be indicative of current mental ill health or with a history of mental ill health.
Follow the instructions carefully.}

\textbf{User:} \emph{Review the following incident report:}
\emph{<report>\{report\}</report>}

\emph{If there are any indicators that any individual in the report is experiencing mental health
issues, highlight quotes from the text containing these indicators (no further discussion
required). Don't respond with more than two quotes. If there are no indicators simply respond with
``EVIDENCE: NONE''.}}

Where evidence is identified, the conversation continues to a second step. The full conversation
history - system prompt, initial query, and model response - is retained in context, and the model
is asked to assess the highlighted evidence according to a three-level scale.

\vulnprompt{%
\textbf{User:} \emph{Based on the evidence highlighted, assess the degree to which it indicates
that an individual is experiencing mental ill health or has a history of mental ill health.
Respond with a short statement justifying your classification, followed by your final
classification, according to the following scale:}

\emph{``EVIDENCE: STRONG'' if ANY of the evidence directly discusses mental ill health, including
treatments or medications for mental ill health or mental health conditions. Note that only one
direct statement is required.}

\emph{``EVIDENCE: MODERATE'' if, in the absence of any direct statements, the most likely
explanation for the accumulated evidence is that an individual is experiencing or has a history of
mental ill health.}

\emph{``EVIDENCE: WEAK'' if there are other, better explanations for the evidence.}}

The STRONG label maps to cases where the text explicitly confirms the vulnerability. MODERATE
captures cases where the evidence is suggestive but not conclusive. WEAK indicates that
alternative explanations are more plausible. Together with NONE (assigned at Step 1), this
produces a four-level ordinal scale per vulnerability type.

The prompts for the remaining three vulnerability types share this structure but include additional
clarifying instructions, developed iteratively during testing to correct systematic
misclassifications by the model. For example, the substance misuse prompt explicitly excludes
recreational drug use and drug possession or distribution that is not indicative of active use,
and the alcohol dependence prompt distinguishes social drinking from problematic intoxication. The
full set of prompts is provided in Appendix~\ref{app:prompts}.

It is worth noting that prompt development of this kind is inherently iterative and, to a degree,
dataset and model specific. The clarifications required here emerged from the particular patterns
of misclassification produced by this model on this corpus; different models, or the same model
applied to incident data from a different force or recorded in a different style, may exhibit
different failure modes requiring different corrective instructions. This has implications for the
generalisability of the prompt set: while the prompts in Appendix~\ref{app:prompts} are provided
in full to support transparency and replication, they should be treated as a starting point rather
than a transferable instrument. Researchers applying similar methods in new contexts should
anticipate a period of iterative testing and prompt refinement as a necessary, and non-trivial,
component of the analytical workflow.

\subsection{Knowledge Distillation}

The computational constraints of the trusted research environment (a single NVIDIA Tesla T4 with
16GB VRAM) limited us to models of approximately 12 billion parameters or fewer. We focused on the
Llama 3 family, which had performed well relative to comparably sized models in the previous study
and for which we had established baseline performance across the instruction-following and
classification tasks relevant to this work. Based on that prior experience and our own
experimentation with the two-step prompts, these models struggled to reliably follow the
instruction set `out of the box'. We therefore used knowledge distillation to train LoRA
adapters \citep{hu2022} for the 8B, 3B and 1B Llama 3 models.

A LoRA adapter is a lightweight addition to an existing language model that modifies its behaviour
for a specific task by training on examples of the desired output - in this case, correctly
formatted two-step classifications. Rather than retraining the full model from scratch, which
would be computationally prohibitive, the adapter introduces a small number of additional
parameters that steer the model's responses while leaving its underlying weights unchanged. The
result is a model that retains the general capabilities of the base Llama 3 model but has learned,
from examples, to follow the structured classification instructions our approach requires.

To train the adapters we generated 10,000 training examples by running GPT-4o-mini through the
two-step classification on a purposive subsample of the Boston FIO dataset used in our previous
study: 2,000 examples showing evidence of each of the four vulnerabilities (8,000 in total), plus a
further 2,000 examples showing no evidence of any vulnerability. Each example captured the full
classification exchange, including
both steps of the process where relevant, so that the adapter would learn to replicate the
complete reasoning pattern, not just isolated responses.

Two technical decisions in the training process are worth explaining. First, the training examples
were generated by a more capable external model (GPT-4o-mini) and used to teach the smaller local
models to reproduce that behaviour, a technique known as knowledge distillation
\citep{kim2016}, in which a ``teacher'' model guides the learning of a smaller ``student'' model. Second, during training, the
model was shown only its own responses as the target for learning, with the classification
instructions themselves excluded from this process. This masking step was necessary because the
same instructions appear in every training example: if the model were allowed to treat the
instructions as part of its target output, it would come to associate good performance with simply
reproducing the instructions rather than with making accurate classifications. By masking them
out, we ensured the model learned the behaviour, i.e.\ how to classify, rather than the scaffolding
around it.

We tested different model sizes and training setups, varying the amount of training data, how many
times the model saw that data, and the LoRA rank (which controls how much capacity the adapter has
to learn new behaviour). We evaluated each configuration against a challenging held-out test set
designed to stress-test the model on difficult, diverse cases. The 8B model substantially
outperformed the smaller variants, and the accuracy gain justified the additional computation cost.
The final configuration used 4-bit quantisation (NF4 with double quantisation;
\citealp{dettmers2023}), a LoRA adapter with
rank 16 and alpha 32 applied to the attention projection layers, trained for a single epoch over
the 10k training examples.

\subsection{Classification Procedure}

Each text segment was classified five times per vulnerability type at a temperature of 0.7 (a
parameter that controls randomness in the model's outputs, discussed further below), producing 20
labels per segment (5 iterations $\times$ 4 vulnerabilities). Multiple classifications per segment
were used to capture label variability across runs.

% =================================================================== RESULTS
\section{Results}

\subsection{Corpus Overview}

The cleaned corpus comprised 2,977 incident logs, producing 3,850 text segments after splitting
longer narratives at the 400-word threshold. Original incident log word counts were heavily
right-skewed (median 181 words, IQR 93--342), and the majority required no splitting: the median
number of segments per log was 1 (IQR 1--1), though the longest log (5,273 words) produced 14
segments. After splitting, segment word counts ranged from 1 to 400 (median 189, IQR 94--372).
Each segment was classified five times per vulnerability type, yielding 77,000 individual
classification events across the four vulnerabilities. Table~\ref{tab:corpus} summarises the
corpus characteristics.

\begin{table}[htbp]
\centering
\small
\begin{tabular}{@{}lr@{}}
\toprule
\multicolumn{2}{@{}l}{\textit{Segmentation}}\\
\midrule
Incident logs & 2,977\\
Incident logs $>$ 400 words & 576\\
Text segments (post-splitting) & 3,850\\
Segments per incident log, median (IQR) & 1 (1--1)\\
Segments per incident log, range & 1--14\\
\midrule
\multicolumn{2}{@{}l}{\textit{Word counts}}\\
\midrule
Incident log word count (pre-split), median (IQR) & 181 (93--342)\\
Incident log word count (pre-split), mean & 277.0\\
Incident log word count (pre-split), range & 1--5,273\\
Segment word count, median (IQR) & 189 (94--372)\\
Segment word count, range & 1--400\\
\midrule
Classification events (segments $\times$ 5 runs $\times$ 4 types) & 77,000\\
\bottomrule
\end{tabular}

\caption{Corpus summary statistics. The upper panel describes the segmentation of incident logs;
the middle panel summarises the word count distribution before and after splitting at the 400-word
threshold; the bottom row reports the total number of individual classifications produced by the
pipeline.}
\label{tab:corpus}
\end{table}

\subsection{Label Variability}

Before examining the accuracy of the classifications, we first assessed how stable the model's
labelling was across repeated runs. Each segment was classified five times at non-zero (0.7)
temperature to distinguish cases where the model is genuinely confident from those where it is
near a decision boundary: if the model's internal probability distribution is centred around a
specific response, moderate temperature will not affect the output, but where it is uncertain,
different runs will land on different labels. Understanding the extent and pattern of this
uncertainty and output variability is important for interpreting any downstream results.

\subsubsection{Consensus}

\begin{figure}[htbp]
\centering
\includegraphics[width=\textwidth]{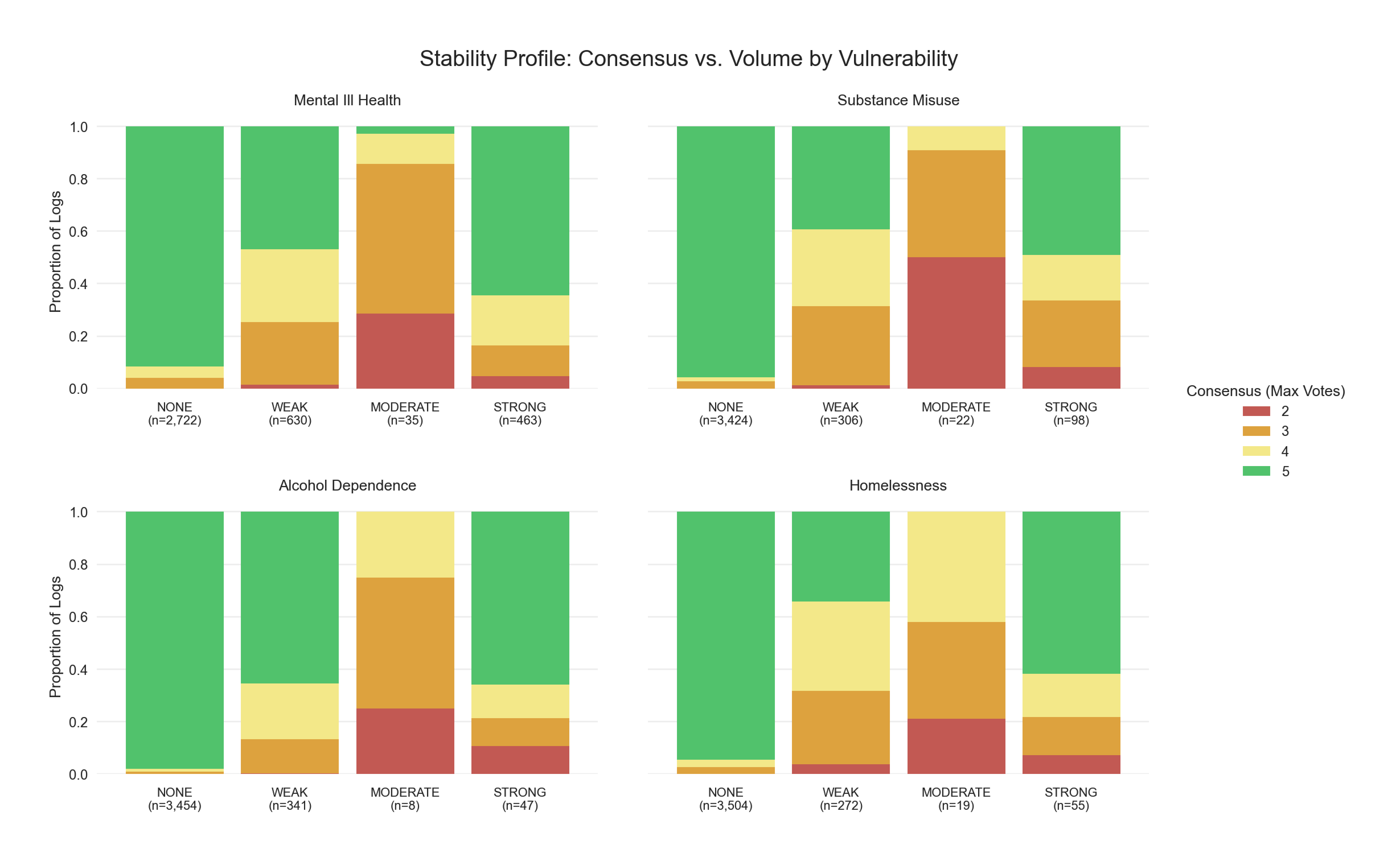}
\caption{Classification stability by vulnerability type and majority label. Stacked bar charts
showing the proportion of text segments at each consensus level (2--5 votes) for each majority
label (NONE, WEAK, MODERATE, STRONG), faceted by vulnerability type. Higher consensus (green)
indicates more stable labelling across runs; lower consensus (red) indicates greater variability.
Bars are annotated with the total number of segments assigned each majority label.}
\label{fig:stability}
\end{figure}

For each text segment, we assigned a majority label by simple plurality across the five runs, with
ties broken towards the stronger evidence category. We then recorded the consensus level - the
number of runs agreeing with the majority label - as a measure of classification stability.
Figure~\ref{fig:stability} shows the distribution of consensus levels, stratified by vulnerability
type and majority label.

Across all four vulnerability types, the NONE label dominated and was highly stable - the vast
majority of segments classified as having no evidence received that label unanimously across all
five runs. Where the model did identify evidence, labelling was considerably less stable. STRONG
labels showed moderate stability, with roughly half to two-thirds of segments reaching full
consensus depending on the vulnerability. WEAK labels showed less extreme disagreement than STRONG
but a lower proportion of unanimous decisions. MODERATE was both the least frequently assigned
label ($n = 84$ across all vulnerabilities combined, compared to 13,104 for NONE, 1,549 for WEAK,
and 663 for STRONG) and was the most variable.

There were modest differences between vulnerability types, with substance misuse classifications
showing slightly more instability than alcohol dependence or mental ill health, but these
differences were not dramatic.

\subsubsection{Severity of Disagreement}

\begin{figure}[htbp]
\centering
\includegraphics[width=\textwidth]{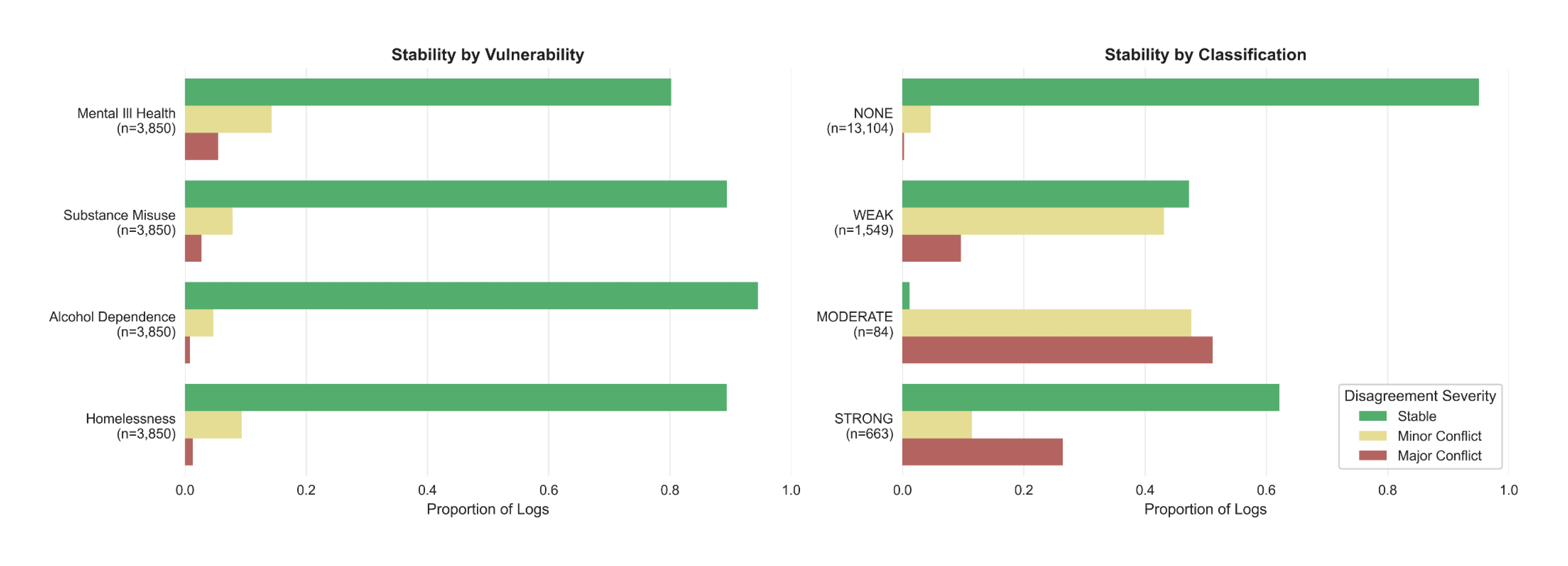}
\caption{Severity of disagreement, stratified by vulnerability type (left) and by majority label
(right). Horizontal bar charts showing the proportion of text segments classified as Stable, Minor
Conflict, or Major Conflict. The left panel facets by vulnerability type ($n = 3{,}850$ segments
per vulnerability). The right panel facets by majority label pooled across all vulnerabilities
(NONE $n = 13{,}104$; WEAK $n = 1{,}549$; MODERATE $n = 84$; STRONG $n = 663$).}
\label{fig:disagreement}
\end{figure}

To better describe the nature of this variability, we categorised each segment's set of five
labels by the severity of disagreement. A segment was classified as \emph{Stable} if all five
labels were identical, \emph{Minor Conflict} if the labels spanned one step on the ordinal scale
(i.e.\ NONE and WEAK, WEAK and MODERATE, or MODERATE and STRONG), and \emph{Major Conflict} if the
labels spanned two or more steps (e.g.\ NONE and STRONG in the same set of five).
Figure~\ref{fig:disagreement} shows the distribution of these categories, stratified by
vulnerability type and by majority label.

By vulnerability type, alcohol dependence classifications were the most stable, followed by
homelessness, then substance misuse, with mental ill health showing the most severe
disagreement - though in each case, these aggregate figures largely reflect the stability of NONE
classifications, which dominate every category. Disaggregating by label reveals a more varied
picture: NONE labels were overwhelmingly stable, STRONG labels showed relatively high rates of
major conflict but low minor conflict and, conversely, WEAK labels showed more minor conflict but
less extreme disagreement. MODERATE, again with very low counts, showed high rates of conflict at
both severities.

The overall picture is that the model is confident and consistent when it finds no evidence, but
where it does identify evidence, it is substantially uncertain about the strength of that evidence
across runs.

\subsection{Accuracy}

\subsubsection{Collapsing the Label Scheme}

The initial four-level classification scheme proved overly granular for the deployed model. The
model could reliably distinguish the presence from the absence of evidence, and could identify
strong evidence with moderate consistency, but was unable to maintain stable distinctions between
weak and moderate evidence across repeated runs. Rather than carry forward a scheme the model
could not reliably operationalise, we collapsed the labelling to reflect the distinctions it could
actually make, defining three categories based on the pattern of votes across the five runs:

\begin{itemize}[leftmargin=1.4em,itemsep=0.15em,topsep=0.4em]
  \item \textbf{No Evidence}: 3 or more NONE votes and fewer than 2 STRONG votes
  \item \textbf{Strong Evidence}: 3 or more STRONG votes and fewer than 2 NONE votes
  \item \textbf{Uncertain}: anything else
\end{itemize}

The first two categories capture segments where the model was consistently confident in one
direction. The ``Uncertain'' category is broader: it includes cases where the model consistently
identified evidence but at sub-strong levels (e.g.\ five WEAK votes), cases where it disagreed with
itself about the strength of evidence across runs, and cases where assessments oscillated between
evidence and absence. In each case, the model judged whatever evidence it found insufficient to
warrant a strong classification. To conduct human review under this collapsed scheme, we developed
a simplified set of labelling heuristics: No Evidence corresponded to segments with no relevant
content or only incidental associations too weak to suggest the vulnerability was present; Strong
Evidence to segments containing confirmatory statements or evidence sufficient to be confident the
vulnerability was present; and Uncertain to segments with suggestive but non-confirmatory evidence.
While this represents a departure from the original four-level scheme, it aligns the labels with
the distinctions the model could reliably make and, we believe, produces the most informative basis
for prevalence estimation.

\subsubsection{Human Review}

We conducted a purposive review of 250 text segments to assess labelling accuracy. The sample was
constructed to ensure coverage across the label distribution: 40 segments that were unanimously
NONE across all four vulnerabilities, approximately 25 segments per vulnerability with 4 or 5
STRONG votes, and the remaining segments from the ``uncertain'' category. Within each stratum,
segments were drawn without replacement using a pseudo-random number generator. Sample sizes per
category were determined by inspecting the
resulting tabulations to ensure sufficient counts for stable estimates of agreement and correction
rates, with additional segments drawn where initial counts were low. Every segment selected for
review on one vulnerability was reviewed for all four, meaning coverage was higher than the
per-vulnerability sampling targets alone would suggest.

\begin{figure}[htbp]
\centering
\includegraphics[width=0.94\textwidth]{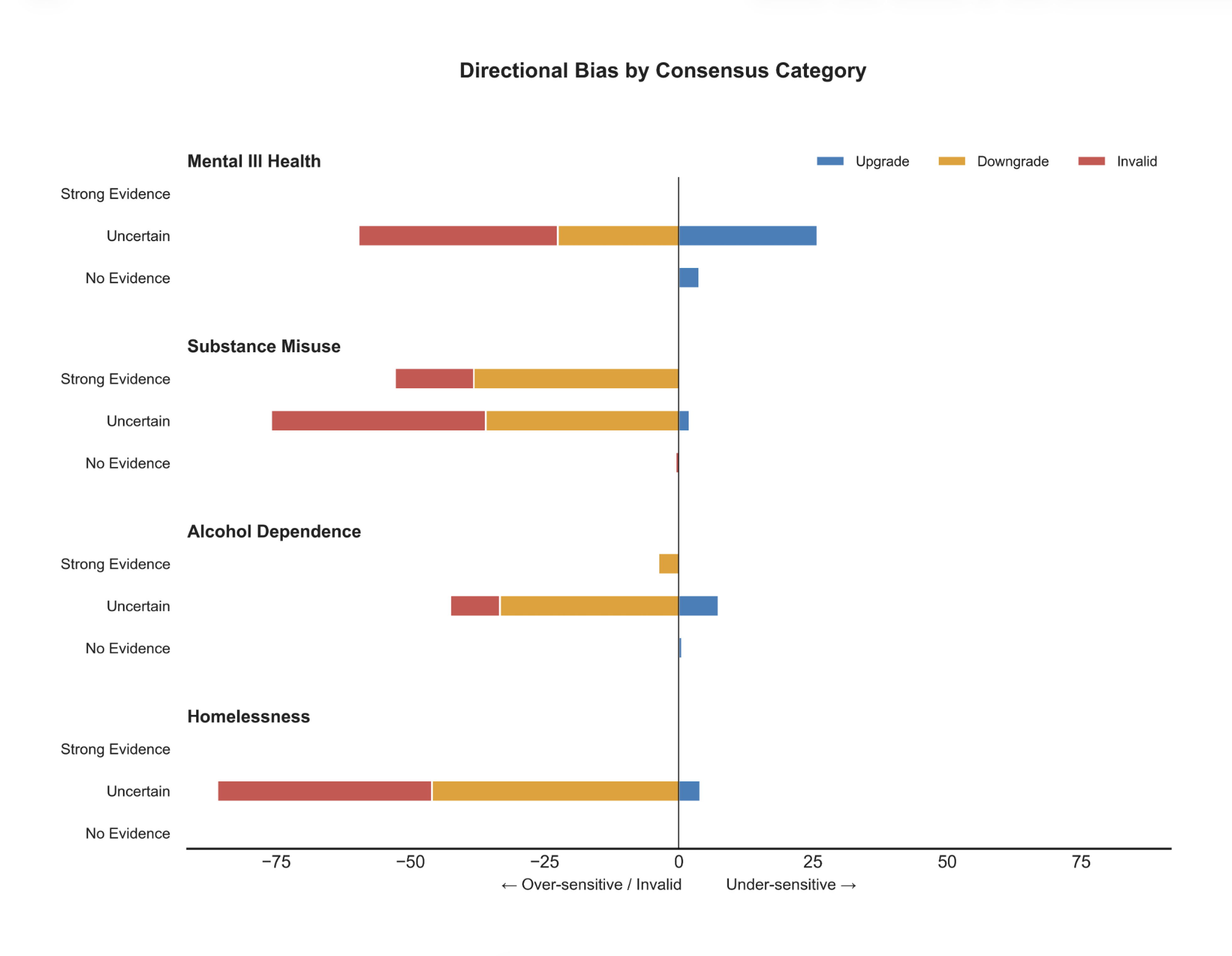}
\caption{Directional bias of model classifications by consensus category. Diverging horizontal bar
chart showing the proportion of reviewed segments that were upgraded (right, indicating
under-sensitivity) or downgraded/marked invalid (left, indicating over-sensitivity), stratified by
vulnerability type and collapsed label. The centre line represents agreement between the model and
human review.}
\label{fig:bias}
\end{figure}

Each segment was reviewed against the model's collapsed label and assigned one of four codes:
\emph{Agree} (the collapsed label is correct), \emph{Upgrade} (the label should be stronger than
assigned), \emph{Downgrade} (there is relevant evidence but it does not warrant the assigned
strength), or \emph{Invalid} (the highlighted evidence is not relevant to the vulnerability in
question). Figure~\ref{fig:bias} and Table~\ref{tab:review} present the results.

\begin{table}[htbp]
\centering
\small
\begin{tabular}{@{}lrrrr@{}}
\toprule
& \multicolumn{3}{c}{LLM classification} & \\
\cmidrule(lr){2-4}
Human review & No Evidence & Uncertain & Strong Evidence & Total\\
\midrule
\multicolumn{5}{@{}l}{\textit{Mental ill health}}\\
Agree      & 126 & 9  & 57 & \textbf{192}\\
Downgrade  & 0   & 14 & 0  & \textbf{14}\\
Upgrade    & 5   & 16 & 0  & \textbf{21}\\
Invalid    & 0   & 23 & 0  & \textbf{23}\\
\textbf{Total} & \textbf{131} & \textbf{62} & \textbf{57} & \textbf{250}\\
\midrule
\multicolumn{5}{@{}l}{\textit{Substance misuse}}\\
Agree      & 165 & 11 & 16 & \textbf{192}\\
Downgrade  & 0   & 18 & 13 & \textbf{31}\\
Upgrade    & 0   & 1  & 0  & \textbf{1}\\
Invalid    & 1   & 20 & 5  & \textbf{26}\\
\textbf{Total} & \textbf{166} & \textbf{50} & \textbf{34} & \textbf{250}\\
\midrule
\multicolumn{5}{@{}l}{\textit{Alcohol dependence}}\\
Agree      & 169 & 27 & 25 & \textbf{221}\\
Downgrade  & 0   & 18 & 1  & \textbf{19}\\
Upgrade    & 1   & 4  & 0  & \textbf{5}\\
Invalid    & 0   & 5  & 0  & \textbf{5}\\
\textbf{Total} & \textbf{170} & \textbf{54} & \textbf{26} & \textbf{250}\\
\midrule
\multicolumn{5}{@{}l}{\textit{Homelessness}}\\
Agree      & 171 & 5  & 29 & \textbf{205}\\
Downgrade  & 0   & 23 & 0  & \textbf{23}\\
Upgrade    & 0   & 2  & 0  & \textbf{2}\\
Invalid    & 0   & 20 & 0  & \textbf{20}\\
\textbf{Total} & \textbf{171} & \textbf{50} & \textbf{29} & \textbf{250}\\
\bottomrule
\end{tabular}

\caption{Cross-tabulation of model labels and human review codes, by vulnerability type. For each
vulnerability, rows show human review codes (Agree, Downgrade, Upgrade, Invalid) and columns show
the model's collapsed label (No Evidence, Uncertain, Strong Evidence), with cell counts and
row/column totals.}
\label{tab:review}
\end{table}

No Evidence labels were highly accurate across all four vulnerabilities, with agreement rates above
96\% (126/131 for mental ill health, 165/166 for drugs, 169/170 for alcohol, 171/171 for
homelessness). The small number of upgrades for mental ill health (5 cases) suggests a modest
false-negative issue for this vulnerability specifically.

Strong Evidence labels were accurate for mental ill health (57/57 agreement), alcohol (25/26), and
homelessness (29/29), but substantially less so for substance misuse, where 13 of 34 strong labels
were downgraded and a further 5 were marked invalid. On review, these errors followed clear
patterns: single instances of drug use being assumed as indicative of longer-term abuse without
sufficient evidence, and persistent confusion between illegal drug use and alcohol consumption.

Uncertain labels showed the most noise across all vulnerabilities. Invalid classifications were
common, often following recurring patterns. Examples include: hospital visits being interpreted as
evidence of mental ill health without supporting context, alcohol consumption being misclassified
as drug-related, and any mention of a person being away from home (staying in a hotel, hospital, or
being reported missing) being flagged as homelessness. The predominant direction of error was
over-sensitivity: across substance misuse, alcohol dependence, and homelessness, the majority of
corrections were downgrades or invalid markings rather than upgrades. Mental ill health again
showed a more mixed pattern, with a notable proportion of upgrades alongside the downgrades.

\subsection{Prevalence Estimates}

To produce incident-level prevalence estimates, we first aggregated from segments back to
incidents. Where an incident comprised multiple segments, the incident was assigned the highest
evidence level found in any of its segments. This reflects the judgement that the presence of
evidence in any part of a narrative is sufficient to flag the incident.

\begin{figure}[htbp]
\centering
\includegraphics[width=0.95\textwidth]{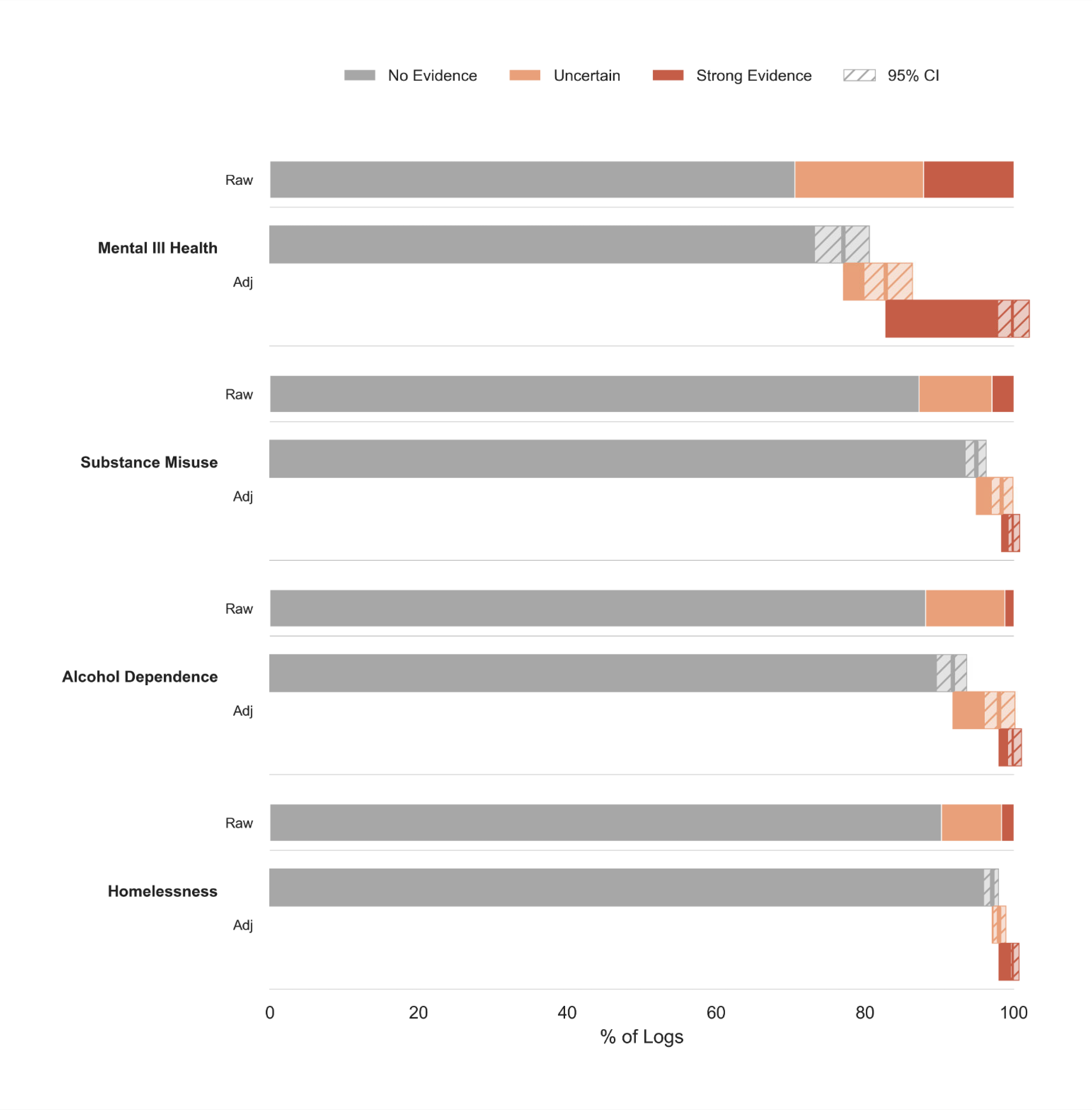}
\caption{Raw and adjusted prevalence estimates by vulnerability type. Stacked horizontal bar chart
comparing raw model classifications (top bar per vulnerability) with bootstrap-adjusted estimates
(cascading bars below). For adjusted estimates, the solid region represents the lower bound of the
95\% CI, the hatched region extends to the upper bound, and a vertical line marks the median.
Panels are arranged vertically by vulnerability type.}
\label{fig:prevalence}
\end{figure}

We then used the human review data to adjust for known classification errors via Monte Carlo
simulation, broadly mirroring established approaches to misclassification correction in
epidemiology \citep{fox2021} and recent methods for valid statistical inference from imperfect
machine-generated labels \citep{angelopoulos2023}. In each of 10,000 iterations, we drew a
bootstrap sample (with replacement) from the 250 reviewed segments, calculated transition
probabilities from each model label to each possible corrected label based on the human review
codes, applied those transition probabilities to relabel all 3,850 segments, and aggregated to the
incident level. This produced a distribution of adjusted prevalence estimates from which we report
the median and 95\% confidence intervals. Figure~\ref{fig:prevalence} and
Table~\ref{tab:prevalence} present both the raw and adjusted estimates.

\begin{table}[htbp]
\centering
\small
\begin{tabular}{@{}lrrrrl@{}}
\toprule
& \multicolumn{2}{c}{Raw (model)} & \multicolumn{3}{c}{Adjusted (Monte Carlo)}\\
\cmidrule(lr){2-3}\cmidrule(lr){4-6}
Category & $N$ & \% & Median $N$ & \% & 95\% CI\\
\midrule
\multicolumn{6}{@{}l}{\textit{Mental ill health}}\\
No Evidence     & 2,101 & 70.6\% & 2,295 & 77.1\% & 73.3--80.6\%\\
Uncertain       & 515   & 17.3\% & 169   & 5.7\%  & 2.8--9.3\%\\
Strong Evidence & 361   & 12.1\% & 510   & 17.1\% & 15.1--19.3\%\\
\textbf{Any Evidence} & \textbf{876} & \textbf{29.4\%} & \textbf{682} & \textbf{22.9\%} & \textbf{19.4--26.7\%}\\
\midrule
\multicolumn{6}{@{}l}{\textit{Substance misuse}}\\
No Evidence     & 2,597 & 87.2\% & 2,827 & 95.0\% & 93.5--96.3\%\\
Uncertain       & 292   & 9.8\%  & 101   & 3.4\%  & 2.1--4.9\%\\
Strong Evidence & 88    & 3.0\%  & 48    & 1.6\%  & 0.9--2.4\%\\
\textbf{Any Evidence} & \textbf{380} & \textbf{12.8\%} & \textbf{150} & \textbf{5.0\%} & \textbf{3.7--6.5\%}\\
\midrule
\multicolumn{6}{@{}l}{\textit{Alcohol dependence}}\\
No Evidence     & 2,624 & 88.1\% & 2,733 & 91.8\% & 89.7--93.7\%\\
Uncertain       & 317   & 10.6\% & 184   & 6.2\%  & 4.3--8.3\%\\
Strong Evidence & 36    & 1.2\%  & 59    & 2.0\%  & 1.2--3.0\%\\
\textbf{Any Evidence} & \textbf{353} & \textbf{11.9\%} & \textbf{244} & \textbf{8.2\%} & \textbf{6.3--10.3\%}\\
\midrule
\multicolumn{6}{@{}l}{\textit{Homelessness}}\\
No Evidence     & 2,687 & 90.3\% & 2,890 & 97.1\% & 96.0--97.9\%\\
Uncertain       & 241   & 8.1\%  & 27    & 0.9\%  & 0.2--1.8\%\\
Strong Evidence & 49    & 1.6\%  & 59    & 2.0\%  & 1.6--2.7\%\\
\textbf{Any Evidence} & \textbf{290} & \textbf{9.7\%} & \textbf{87} & \textbf{2.9\%} & \textbf{2.1--4.0\%}\\
\bottomrule
\end{tabular}

\caption{Raw and adjusted prevalence estimates by vulnerability type ($N = 2{,}977$ incident logs).
For each vulnerability, the table shows the raw count and percentage of incidents classified as No
Evidence, Uncertain, and Strong Evidence, alongside the bootstrap-adjusted median count and
percentage with 95\% confidence intervals. An ``Any Evidence'' row combines Uncertain and Strong.}
\label{tab:prevalence}
\end{table}

The adjustment had a substantial effect across all vulnerability types, consistently reducing the
estimated prevalence of statements that suggest the presence of vulnerabilities, in line with the
over-sensitivity identified in the error analysis above. For mental ill health, raw estimates
suggested 29.4\% of incidents contained some evidence of vulnerability; after adjustment, this fell
to 22.9\% (95\% CI: 19.4--26.7\%). The adjustment was particularly notable in the Uncertain
category, which shrank from 17.3\% to 5.7\%, consistent with the high rate of over-sensitive
labelling found in the human review. Conversely, the Strong Evidence estimate for mental ill health
actually increased after adjustment (from 12.1\% to 17.1\%), reflecting the upgrade
corrections identified during review.

Substance misuse showed the most dramatic correction: raw estimates of 12.8\% with any evidence
were adjusted down to 5.0\% (95\% CI: 3.7--6.5\%), driven by the high rate of invalid and
downgraded classifications found during review. Alcohol dependence estimates reduced from 11.9\% to
8.2\% (95\% CI: 6.3--10.3\%), and homelessness from 9.7\% to 2.9\% (95\% CI: 2.1--4.0\%).

Across all four vulnerabilities, the pattern was consistent: the model's raw output over-estimates
the prevalence of vulnerability indicators, primarily because its uncertain classifications contain
a high proportion of over-sensitive or invalid flags. The adjustment procedure brings estimates
closer to what we would expect based on human judgement, though the confidence
intervals - particularly for the less prevalent vulnerabilities - remain wide, reflecting both the
inherent uncertainty in the model's labelling and the limited size of the human review sample.

\subsection{Comparison with Force Qualifiers}

Our final analysis compares the model's classifications against an additional administrative
indicator available within the data. The incident logs included linked categorical qualifiers
assigned by force control room operators at the point of recording, covering mental ill health,
substance misuse, and alcohol dependence. These offered an apparent external benchmark against
which to assess the model's classifications for the three vulnerability types where both sources
were available. Table~\ref{tab:qualifiers} cross-tabulates the force qualifiers against the model's
collapsed labels for the three vulnerabilities where both sources were available.

\begin{table}[htbp]
\centering
\small
\begin{tabular}{@{}lrrrr@{}}
\toprule
& \multicolumn{3}{c}{LLM classification} & \\
\cmidrule(lr){2-4}
Force qualifier & No Evidence & Uncertain & Strong Evidence & Total\\
\midrule
\multicolumn{5}{@{}l}{\textit{Mental ill health}}\\
Qualifier absent  & 1,980 & 419 & 220 & \textbf{2,619}\\
Qualifier present & 121   & 96  & 141 & \textbf{358}\\
\textbf{Total}    & \textbf{2,101} & \textbf{515} & \textbf{361} & \textbf{2,977}\\
\midrule
\multicolumn{5}{@{}l}{\textit{Substance misuse}}\\
Qualifier absent  & 2,586 & 276 & 81 & \textbf{2,943}\\
Qualifier present & 11    & 16  & 7  & \textbf{34}\\
\textbf{Total}    & \textbf{2,597} & \textbf{292} & \textbf{88} & \textbf{2,977}\\
\midrule
\multicolumn{5}{@{}l}{\textit{Alcohol dependence}}\\
Qualifier absent  & 2,594 & 267 & 34 & \textbf{2,895}\\
Qualifier present & 30    & 50  & 2  & \textbf{82}\\
\textbf{Total}    & \textbf{2,624} & \textbf{317} & \textbf{36} & \textbf{2,977}\\
\bottomrule
\end{tabular}

\caption{Cross-tabulation of force qualifier flags and model classifications by vulnerability type.
Rows indicate whether the force's qualifier was present or absent; columns show the model's
collapsed label.}
\label{tab:qualifiers}
\end{table}

Across all three vulnerabilities, the qualifiers and model labels showed substantial disagreement.
For mental ill health, 121 of 358 qualifier-present incidents (33.8\%) showed no narrative evidence
according to the model, while 220 incidents with strong narrative evidence carried no qualifier.
The drug and alcohol qualifiers were applied far less frequently - 34 and 82 incidents
respectively - but the pattern of disagreement within those smaller numbers was no less pronounced.
Of 34 drug-flagged incidents, only 7 showed strong narrative evidence of substance misuse and 11
showed none at all. For alcohol, just 2 of 82 flagged incidents showed strong evidence, and 30
showed none. In each case, the qualifiers picked up a largely different set of incidents to those
identified by the model.

Careful but non-systematic review of the underlying narratives suggested that this disagreement was
not simply model error, or a matter of borderline cases falling on different sides of a threshold.
Across all three vulnerabilities, qualifier-present incidents with no model evidence were
frequently mundane - prank calls, road traffic incidents, dropped calls - with no discernible
content relating to the flagged vulnerability. Qualifier-absent incidents flagged by the model, by
contrast, contained explicit references to the vulnerability in question: named mental ill health
conditions, suicidality and self-harm, visible intoxication, or direct statements of drug use. The
qualifiers, in many cases, did not appear to correspond to anything in the narrative at all.

Two broad interpretations of these discrepancies are plausible. First, where a qualifier is present
but there is no corresponding evidence in the narrative, the flag may be drawing on information
that is not captured in the incident log itself. This could include intelligence from other
systems, prior contact histories, or details carried forward from earlier incidents linked to the
same individual or address. In this sense, the qualifier may be meaningful within an operational
context but not observable in the data available for analysis. More generally, these administrative
datasets were not created for research purposes, and it is difficult to fully establish what they
represent from the extracted fields alone. This introduces uncertainty and suggests that caution is
needed when interpreting them.

Second, and potentially more concerning, are cases where the narrative contains clear references to
a vulnerability, such as specific mental health conditions, but no corresponding qualifier is
present. While it remains possible that additional information not available to us influences
whether a qualifier is applied (or omitted), discussions with a range of operational policing
partners point to a more practical explanation. In fast-moving and pressurised call-handling
environments, the use of qualifiers is not always consistent, and their presence or absence is
unlikely to affect the immediate response or support provided. Qualifiers may therefore reflect an
administrative process that is applied unevenly rather than a reliable record of incident content.

Importantly, taken together, these findings indicate that the qualifier fields do not provide a
sufficiently robust basis for assessing model performance. They appear to capture something
different from the narrative evidence, and only partially overlap with it. For this reason, we do
not treat them as a form of ground truth in this analysis. More broadly, the inconsistencies
observed here suggest that analyses which rely on these qualifiers should be interpreted with care,
and they highlight the value of approaches such as the one explored here that work directly with
the narrative data.

% ================================================================ DISCUSSION
\section{Discussion}

This study set out to test whether an LLM-based classification approach, initially developed and
validated on open-source police data from the US \citepalias{analyzeboston}, could be adapted to produce
meaningful prevalence estimates of vulnerability indicators in UK-based police incident logs. This
was assessed within practical constraints governing access to, storage of, and analysis of
sensitive data - constraints that shaped key aspects of the analytical design, but which are
broadly representative of the data governance conditions that police forces and their partners
routinely face. The findings suggest that such adaptation is possible, but that it requires
considerably greater care than is often implied in prevailing accounts of LLM development.

Our earlier work on the Boston FIO dataset indicated that, in that setting, instruction-tuned LLMs
were effective at identifying the absence of vulnerability indicators within police narratives. The
finding that LLMs are highly effective ``negative filters'' was the most robust result across every
model size and prompting strategy explored. That finding replicates convincingly here. Even after
fine-tuning a distilled small (8B parameter) model, training it on American data, anonymising the
input text, and deploying it on British incident logs, the model's ability to screen out narratives
with no vulnerability content remained strong: `No Evidence' labels were stable across repeated
runs and accurate on human review, with agreement rates exceeding 96\% across all four
vulnerability types.

Beyond replicating the negative filter, this study explored means by which structured human review
of positive and uncertain labels could be used to produce credible prevalence estimates. The
estimates we report - mental ill health indicators in approximately 23\% of incidents, alcohol
dependence in 8\%, substance misuse in 5\%, and homelessness in 3\% - are, to our knowledge, the
first such estimates derived from routine UK police incident narratives through generative AI-based
semantic analysis. These are not numbers that could feasibly be produced by manual coding at this
scale, nor are they currently captured by existing recording practices within any police force that
we are aware of.

The comparison with the force's own qualifier flags underscores this last point. As reported in the
results, the qualifiers and model classifications picked up largely different sets of incidents, and
the character of that disagreement - mundane narratives carrying flags, explicit vulnerability
content going unflagged - was not consistent with two noisy measures of the same underlying
construct. We are not aware of any mechanism within the force's recording practices that can
currently produce the kind of narrative-level vulnerability identification that this study attempts.
If police forces or researchers want a better understanding of the day-to-day prevalence of these
types of interactions in frontline encounters, administrative flags alone would seem insufficient.
Reading unstructured text gives you a different picture and, we would argue, an informative one.

That said, it is important to reinforce that the prevalence estimates we report are the product of a
multi-stage pipeline: repeated classifications, label aggregation, structured human review, and
Monte Carlo correction. We did not arrive at this design by choice so much as by necessity. Each
stage was introduced because the output of the preceding one was not fit for purpose, and notably,
the stages at which the pipeline fails most seriously are precisely those that operate without human
input.

The simplest conceivable deployment of an LLM classifier - and the one most consistent with industry
messaging around ease of adoption - is to run each narrative through the model once and accept the
label. Our label variability analysis, both in the previous study and this one, shows why this would
be dangerous. Repeated classifications of the same narrative frequently disagree: cases labelled as
showing no evidence on one run are flagged as containing strong evidence on another, and vice versa. The
negative filter, the model's reliable ability to screen out narratives with no vulnerability
content, is a property that emerges from aggregation across runs. It does not describe the behaviour
of any single classification. A researcher or practitioner who ran the model once and accepted the
output would be working with labels that are noisy in both directions: over-flagging some cases,
missing others, with no way to distinguish which errors they were making or how often. There would
be limited basis for knowing how far the resulting estimates were from reality, or in which
direction - only the most obvious errors would be detectable.

Running the model multiple times and aggregating, as we do, is what underpins the negative filter
finding and stabilises the positive end of the label distribution. But it does not solve the
problem. The aggregated labels still substantially overestimate every vulnerability we sought to
identify. The model systematically flags evidence where a human reviewer finds none - hotel stays
interpreted as homelessness, hospital visits as mental ill health, alcohol consumption as substance
misuse. Some of these errors are obvious on inspection; others are plausible enough to survive
casual scrutiny. Crucially, the pattern is not detectable from the model's output alone. Without
structured human review across a large and carefully sampled set of cases, a user would have no
basis for knowing that the over-sensitivity exists \citep{pangakis2025}, let alone for estimating
its scale. The Boston
study demonstrated the same pattern - reliable negative filtering alongside systematic
over-sensitivity - across multiple off-the-shelf models and prompting strategies, and the fact that
it persists here in a fine-tuned, purpose-built model suggests this may be a property of the task,
not just an artefact of insufficient engineering. At the very least, it suggests that all LLM
classification tasks should be subject to careful human review.

The Monte Carlo adjustment we apply is an example of the methodological rigour required to correct
for the over-sensitivity we detected. Human review of a substantial portion of the narratives was
required to estimate transition probabilities from model labels to corrected labels, and propagate
that correction across the full corpus via simulation. The resulting confidence intervals are wide,
particularly for less prevalent vulnerabilities, reflecting both the ambiguity of the task and the
limitations of the model. The scale of the correction is not trivial: the raw ``any evidence'' rate
for substance misuse was 12.8\%; the adjusted estimate was 5.0\%. These are discrepancies that could
materially distort operational priorities or resource allocation if taken at face value.

The implication is that producing trustworthy population-level estimates from LLM-based
classification required repeated runs, domain-informed human review, and statistical
correction - precisely the kinds of methodological infrastructure that LLMs are often presented as
streamlining, reducing, or in some cases replacing. Nothing in our experience, albeit across only
two studies, but with different data, different models, and different deployment contexts, provides
support for that narrative. The model is a powerful component, but the expertise required to use it
responsibly is no less than would be expected of any other empirical measurement instrument, and
arguably more, because the failure modes are both less visible and may be amplified by the speed and
scale at which such systems operate. A poorly calibrated survey or a biased sampling frame will be
familiar to most quantitative researchers. By contrast, a model that produces fluent, confident, and
systematically incorrect classifications is a different class of problem, and one for which the
default mode of deployment offers little in the way of diagnostic signals.

None of this should be taken as grounds for dismissing LLM-based classification. The estimates we
produce could not feasibly have been generated by manual coding at this scale, and they reveal
patterns of vulnerability that are invisible to existing recording practices. But at the population
level, they required substantial correction to be defensible, and at the individual level, they
should not be trusted. Individual classifications are frequently wrong - over-sensitive,
inconsistent, and sometimes based on evidence that was not relevant to the vulnerability in
question. Our adjustment works because these errors are systematic enough to estimate and correct in
the aggregate; at the level of a single label, no such correction is possible. Where model outputs
might be used to triage, flag, or prioritise specific cases - as is increasingly being proposed in
some policing contexts - the margin for error we document here argues strongly against it, whether
labels are drawn from a single run or aggregated across many. Establishing whether individual labels
could suitably support such decisions, and how this might be approached responsibly, is a topic that warrants further empirical study
that is beyond the scope of this study.

Beyond these broader issues, there are several specific limitations to this study that should be
acknowledged. First, the prevalence adjustment rests on human review by a single reviewer without a
formal inter-rater reliability exercise. This limitation is shared by any prevalence estimation
derived from interpretive coding at scale. Manual qualitative coding of the full corpus would face
the same constraint, and the resource intensity of such an exercise is precisely what motivates the
approach. That said, the review functioned as an error audit of model output rather than a
traditional content analysis, and the corrections driving the largest adjustments were not
borderline interpretive judgements. Many of the Invalid classifications, i.e.\ hotel stays flagged as
homelessness, or hospital visits as mental ill health, are errors any competent reviewer would
catch. The cases where reviewer subjectivity genuinely matters are the Downgrade calls within the
Uncertain category, which contribute a smaller share of the overall correction. The corrections were
also overwhelmingly unidirectional: the model over-assigns vulnerability indicators relative to the
narrative evidence, and it is unlikely that a different reviewer would reach a materially different
conclusion. A more formal exercise would refine the estimates, but we would not expect it to alter
the overall pattern. Relatedly, the Uncertain category likely contains heterogeneous error profiles
that our review was not designed to distinguish, though the practical value of stratifying further
is limited given that the category was defined precisely by the model's inability to resolve these
cases at finer resolution.

Second, this study tested a single classification pipeline: one fine-tuned model, one prompt design,
one set of parameters. The design space for LLM-based classification is vast, and we cannot claim to
have identified an optimal configuration. A comprehensive comparison across models, prompting
strategies, and fine-tuning approaches is neither practical nor the purpose of this study. The
pipeline was designed with care and informed by prior experimental work on comparable data. More
importantly, the broader patterns we observe - strong negative filtering, systematic
over-sensitivity on ambiguous cases, and the need for human review to produce trustworthy
estimates - are consistent with our earlier findings across multiple model sizes and prompting
strategies. Again, this may suggest that these patterns are not specific to our configuration but
instead reflect more general characteristics of applying LLMs to this class of problem, although
further research would be needed to establish whether this holds across other datasets, models, and
deployment contexts.

Third, incident narratives were split into segments at a 400 word boundary, and each segment was
classified without reference to the rest of the incident. Because an incident is assigned the highest
evidence level found in any of its segments, splitting might suppress an indicator whose evidence is
distributed across a narrative rather than stated in any single segment. However, our experience with
the data suggests that indicators are typically recorded as brief factual statements - a reference to
a mental health team, an observation of intoxication, a note of no fixed abode - rather than
distributed over a long account, and 80.65\% of logs were 400 words or less and were classified whole.
We suspect that the effect of splitting is to increase sensitivity: a segmented narrative has greater
opportunity to return a positive label under the maximum rule, but we are unable to test this
directly. The threshold was set by the computational constraints of the secure research environment
rather than chosen freely, so re-running at a larger segment size was not possible.

Finally, the data analysed come from a single UK police force and are drawn from a relatively
limited temporal and spatial window. The dataset was anonymised via whitelisting, and the model was
fine-tuned on US police narratives that are adjacent to, but likely differ in important ways from,
logs recorded in UK policing data. The whitelisting may therefore have removed contextual
information that could itself serve as a vulnerability indicator, while the domain transfer from US
to UK policing language is difficult to disentangle from the distillation process. On manual review,
we did not identify any errors that could be attributed to differences between the language in the US
and UK datasets; the comprehension failures we observed were of a kind similar to those we observed
when working with the US corpus. However, we cannot rule out that performance would differ on
non-redacted or more closely domain-matched data.

% ================================================================ CONCLUSION
\section{Conclusion}

This study demonstrates that fine-tuned LLMs can produce meaningful, if imperfect, prevalence
estimates of vulnerability indicators in routine police incident narratives - information that does
not currently exist through any other scalable mechanism. The estimates themselves offer a novel
lens on the burden of vulnerability in day-to-day policing: mental ill health indicators in over one
in every five incidents, with lower but non-trivial rates for alcohol dependence, substance misuse,
and homelessness. These are approximate figures with wide confidence intervals, particularly for the
less prevalent vulnerabilities, and they should be understood as indicative orders of magnitude
rather than precise measurements.

Having examined both the strengths and limitations of this pipeline, we suggest that its primary
value lies in application at greater scale. Extending the approach across multiple years and
force-wide datasets would support more robust demand modelling and provide a clearer evidence base
for decisions around resourcing, workforce development, and multi-agency safeguarding arrangements,
areas where quantitative insight is currently limited. Similarly, applying a consistent approach
across multiple forces may enable more meaningful comparisons in the scale and composition of
vulnerability-related demand, supporting more informed benchmarking and strategic planning across
the policing system.

At the same time, the findings highlight the distance between raw model output and defensible
measurement. While LLMs offer powerful new capabilities for extracting information from unstructured
data at scale, their outputs do not, in themselves, constitute reliable evidence. Producing
estimates that can withstand scrutiny in this context requires a broader methodological framework,
including replication, human oversight, and statistical adjustment. This is not a temporary
limitation of a particular implementation, but reflects more fundamental challenges in applying these
models to complex, interpretive classification tasks. In this sense, we believe the responsible use
of LLMs in sensitive domains depends not only on model performance, but on the rigour of the
processes within which they are embedded.

% ============================================================== DECLARATIONS
\section*{Declarations}
\addcontentsline{toc}{section}{Declarations}
\small

\subsection*{Availability of data and materials}
The classification pipeline, fine-tuning scripts, and analysis code are available at
\url{https://github.com/samrelins/vulnerability_classifier_pipeline}. The 10,000 knowledge
distillation training examples, generated from the publicly available Boston Police Department FIO
dataset, are available at \url{https://osf.io/har9m/overview}. The UK STORM incident logs analysed
in this study cannot be made publicly available; access was granted by a UK police force under data
governance conditions that preclude external sharing.

\subsection*{Competing interests}
The authors have no competing interests to declare that are relevant to the content of this article.

\subsection*{Funding}
The support of the Economic and Social Research Council (ESRC) is gratefully acknowledged. Grant
reference number: ES/W002248/1.

\subsection*{Authors' contributions}
Conceptualisation: SR (equal), DB (equal). Funding acquisition: DB (lead). Methodological design: SR
(equal), DB (equal). Data curation: SR (lead), DB (supporting). Software development: SR (lead).
Formal analysis: SR (lead). Visualisation: SR (lead). Project administration: SR (lead). Supervision:
DB (lead). Writing - original draft: SR (lead), DB (equal). Writing - review and editing: SR (equal),
DB (equal).

\subsection*{Acknowledgements}
The authors would like to thank Charlie Lloyd for his valuable comments on an earlier draft of this
manuscript. They also gratefully acknowledge the police officers and staff at the participating
police force for their support throughout the project.

Claude (Anthropic) and ChatGPT (OpenAI) were used to copy-edit author-written text and to provide
critical review of the manuscript's argument and expression. The argument, analysis, and conclusions
are the authors' own. The authors assessed each suggested edit and review point, decided which to act
on, and take full responsibility for the content of the manuscript.

\normalsize

% =============================================================== REFERENCES
\bibliography{refs}

% ================================================================= APPENDIX
\appendix
\section{Classification Prompts by Vulnerability Type}
\label{app:prompts}

The mental ill health prompts are presented in the main text (Section~\ref{sec:twostep}). The
prompts for the remaining three vulnerability types are reproduced below. All follow the same
two-step structure, with additional clarifying instructions where noted.

\subsection*{Substance misuse}

\vulnprompt{%
\textbf{System prompt.}

\emph{You are an assistant tasked with identifying statements in incident reports that might be
indicative of current substance abuse or a history of substance abuse problems. Note that substance
abuse relates specifically to drugs of abuse, and should not be associated with recreational drugs
like marijuana, alcohol or recreational cocaine use. Follow the given instructions carefully.}

\medskip
\textbf{Step 1 - Evidence extraction prompt.}

\emph{Review the following incident report:}

\emph{<report>\{report\}</report>}

\emph{If there are any indicators that suggest the active use of drugs of abuse, or a
current/history of addiction to drugs of abuse, highlight quotes from the text containing these
indicators (no further discussion required). Don't respond with more than two quotes. If there are
no indicators simply respond with ``EVIDENCE: NONE''.}

\emph{Don't highlight any evidence that points to drug distribution, sale or possession that isn't
otherwise indicative of drug \textbf{use}.}

\medskip
\textbf{Step 2 - Evidence scrutiny prompt.}

\emph{Based on the evidence highlighted, assess the degree to which it indicates that an individual
is actively using drugs of abuse, or has a current addiction to drugs of abuse. Respond with a short
statement justifying your classification, followed by your final classification, according to the
following scale:}

\emph{``EVIDENCE: STRONG'' if ANY of the evidence directly discusses the use of drugs of abuse or
drug addiction, including treatments or services for drug addiction or abuse. Note that only one
direct statement is required.}

\emph{``EVIDENCE: MODERATE'' if, in the absence of any direct discussion, the most likely
explanation for the accumulated evidence is that an individual is currently suffering from drug
addiction or is actively using drugs of abuse.}

\emph{``EVIDENCE: WEAK'' if there are other, better explanations for the evidence.}}

\subsection*{Alcohol dependence}

\vulnprompt{%
\textbf{System prompt.}

\emph{You are an assistant tasked with identifying statements in incident reports that might
indicate an individual is an alcoholic or has a drinking problem. Follow the given instructions
carefully.}

\medskip
\textbf{Step 1 - Evidence extraction prompt.}

\emph{Review the following incident report:}

\emph{<report>\{report\}</report>}

\emph{If there are any indicators that suggest alcoholism or drinking problems, highlight quotes
from the text containing these indicators (no further discussion required). Don't respond with more
than two quotes. If there are no indicators simply respond with ``EVIDENCE: NONE''.}

\emph{Note:}

\emph{-- Do not highlight evidence that simply describes an individual drinking alcohol, or social
drinking that isn't otherwise obviously problematic.}

\emph{-- Do highlight extreme intoxication, or intoxication that has clearly contributed to a
situation where the police have been called to intervene.}

\medskip
\textbf{Step 2 - Evidence scrutiny prompt.}

\emph{Based on the evidence highlighted, assess the degree to which it indicates that an individual
is an alcoholic or has a drinking problem. Respond with a short statement justifying your
classification, followed by your final classification, according to the following scale:}

\emph{``EVIDENCE: STRONG'' if ANY of the evidence directly discusses alcoholism, or services that
are SPECIFICALLY RELATED to alcoholism or sobriety (that isn't drug-related). Note that only one
direct statement is required.}

\emph{``EVIDENCE: MODERATE'' if, in the absence of any direct discussion, the most likely
explanation for the accumulated evidence is that an individual is an alcoholic or has a drinking
problem.}

\emph{``EVIDENCE: WEAK'' if there are other, better explanations for the evidence.}}

\subsection*{Homelessness}

\vulnprompt{%
\textbf{System prompt.}

\emph{You are an assistant tasked with identifying statements in incident reports that are evidence
of homelessness. Follow the given instructions carefully.}

\medskip
\textbf{Step 1 - Evidence extraction prompt.}

\emph{Review the following incident report:}

\emph{<report>\{report\}</report>}

\emph{If there are any indicators that suggest homelessness, highlight quotes from the text
containing these indicators (no further discussion required). Don't respond with more than two
quotes. If there are no indicators simply respond with ``EVIDENCE: NONE''.}

\emph{Note:}

\emph{-- ANY statements that reference homelessness should be flagged as evidence, be they relating
to a location or to individuals not present or not involved in the incident. Flag ALL discussion of
homelessness.}

\medskip
\textbf{Step 2 - Evidence scrutiny prompt.}

\emph{Based on the evidence highlighted, assess the degree to which it indicates that an individual
is currently homeless. Respond with a short statement justifying your classification, followed by
your final classification, according to the following scale:}

\emph{``EVIDENCE: STRONG'' if ANY of the evidence directly discusses homelessness, or services that
are SPECIFICALLY RELATED to homelessness. Note that only one direct statement is required.}

\emph{``EVIDENCE: MODERATE'' if, in the absence of any direct discussion, the most likely
explanation for the accumulated evidence is that an individual is currently homeless.}

\emph{``EVIDENCE: WEAK'' if there are other, better explanations for the evidence.}}

\end{document}